\documentclass{article}

\usepackage{arxiv}

\usepackage[utf8]{inputenc} 
\usepackage[T1]{fontenc}    
\usepackage{hyperref}       
\usepackage{url}            
\usepackage{booktabs}       
\usepackage{amsfonts}       
\usepackage{nicefrac}       
\usepackage{microtype}      
\usepackage{lipsum}		
\usepackage{graphicx}
\usepackage{natbib}
\usepackage{doi}
\usepackage{tabularx}
\usepackage{booktabs}
\usepackage{comment}
\usepackage{amsmath}
\usepackage{xcolor}
\usepackage{float}
\usepackage{listings}
\usepackage{hyperref}

\usepackage{graphicx}   
\usepackage{array} 
\usepackage{multirow}
\usepackage{placeins}

\title{Early Risk Stratification of Dosing Errors in Clinical Trials Using Machine Learning}

\author{ \href{https://orcid.org/0009-0004-3425-8189}{\includegraphics[scale=0.06]{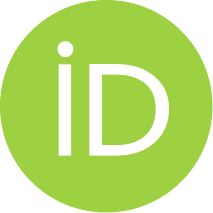}\hspace{1mm}Félicien Hêche}\thanks{These authors contributed equally to this work.} \\
	Department of Radiology and Medical Informatics \\
	University of Geneva \\
	Geneva, Switzerland \\
	\texttt{felicien.heche@unige.ch} \\
    \And
    \href{https://orcid.org/0000-0003-3768-6408}{\includegraphics[scale=0.06]{orcid.pdf}\hspace{1mm}
	Sohrab Ferdowsi}\footnotemark[1]  \\
	Department of Radiology and Medical Informatics \\
	University of Geneva \\
	Geneva, Switzerland \\
	\texttt{sohrab.ferdowsi@unige.ch} \\
	\And
	\href{https://orcid.org/0000-0003-3309-6128}{\includegraphics[scale=0.06]{orcid.pdf}\hspace{1mm} Anthony Yazdani} \\
	Department of Radiology and Medical Informatics \\
	University of Geneva \\
	Geneva, Switzerland \\
	\texttt{anthony.yazdani@unige.ch} \\
    \AND
    Sara Sansaloni-Pastor \\
    Actelion Pharmaceuticals Ltd \\
	  Basel, Switzerland \\
	\texttt{SSansalo@ITS.JNJ.com} \\
	\And
	\href{https://orcid.org/0000-0001-6238-4503}{\includegraphics[scale=0.06]{orcid.pdf}\hspace{1mm} Douglas Teodoro} \\
    Department of Radiology and Medical Informatics \\
	University of Geneva \\
	Geneva, Switzerland \\
	\texttt{douglas.teodoro@unige.ch} \\
}

\renewcommand{\shorttitle}{Early Risk Stratification of Dosing Errors in CTs Using ML}

\begin{document}
\maketitle

\begin{abstract}
\textbf{Objective:} The objective of this study is to develop a machine learning (ML)-based framework for early risk stratification of clinical trials (CTs) according to their likelihood of exhibiting a high rate of dosing errors, using information available prior to trial initiation.  \\
\textbf{Materials and Methods:}  We constructed a dataset from ClinicalTrials.gov comprising 42,112 CTs. Structured, semi-structured trial data, and unstructured protocol-related free-text data were extracted. CTs were assigned binary labels indicating elevated dosing error rate, derived from adverse event reports, MedDRA terminology, and Wilson confidence intervals. We evaluated an XGBoost model trained on structured features, a ClinicalModernBERT model using textual data, and a simple late-fusion model combining both modalities. Post-hoc probability calibration was applied to enable interpretable, trial-level risk stratification. \\
\textbf{Results:} The late-fusion model achieved the highest AUC–ROC (0.862). Beyond discrimination, calibrated outputs enabled robust stratification of CTs into predefined risk categories. The proportion of trials labeled as having an excessively high dosing error rate increased monotonically across higher predicted risk groups and aligned with the corresponding predicted probability ranges. \\
\textbf{Discussion:} These findings indicate that dosing error risk can be anticipated at the trial level using pre-initiation information. Probability calibration was essential for translating model outputs into reliable and interpretable risk categories, while simple multimodal integration yielded performance gains without requiring complex architectures. \\
\textbf{Conclusion:} This study introduces a reproducible and scalable ML framework for early, trial-level risk stratification of CTs at risk of high dosing error rates, supporting proactive, risk-based quality management in clinical research.
\end{abstract}



\section{Introduction} \label{introduction}

Medication errors, defined as failures in the treatment process that lead to, or have the potential to lead to, harm to the patient \cite{aronson2009medication}, constitute a significant threat to public health worldwide. Despite sustained efforts to improve medication safety, these events continue to impose a substantial burden on healthcare systems \cite{elliott2021economic, rouhani2018application, hodkinson2020preventable}. In response, major health authorities have prioritized medication safety as a core component of patient safety strategies. Notably, the World Health Organization (WHO) launched the Medication Without Harm initiative to catalyze coordinated action aimed at reducing severe, avoidable medication-related harm \cite{challenge2017medication}, and has since reinforced this commitment through targeted policy guidance \cite{world2024medication}.

In parallel with these policy-driven efforts, machine learning (ML) approaches have been increasingly adopted in healthcare to support tasks such as medical image interpretation \cite{shen2017deep, carriero2024deep}, early detection of disease \cite{govindu2023early, song2020predictive}, personalized treatment planning \cite{bertsimas2020personalized, tan2022tree}, and the assessment of operational risks in clinical trials (CTs) \cite{teodoro2025scoping, ferdowsi2023deep, ferdowsi2021classification}. A key strength of ML lies in its ability to model high-dimensional, heterogeneous, and often multimodal data \cite{acosta2022multimodal, warner2024multimodal}, enabling the discovery of complex associations between patient characteristics, treatments, and outcomes that may elude conventional analytic approaches \cite{cerqueira2022case, yazdani2025evaluation}. When appropriately integrated into clinical and research workflows, such models can support timely decision-making and contribute to improved medication safety \cite{topol2019high, berge2023machine}.

Despite these advances, important limitations remain in the application of ML to medication error detection. Specifically, as highlighted in a recent scoping review \cite{Heche2026MLErrorScoping}, existing ML-based studies focus almost exclusively on errors occurring during routine clinical care. Notably, this review identified no prior work addressing medication errors arising in research and development settings. This gap is surprising given the importance of the pharmaceutical research sector, which involves annual global investments exceeding US \$250 billion \cite{chandra2024comprehensive}, as well as the potential consequences of such errors. Indeed, medication errors in CTs can compromise trial validity and data integrity \cite{vrijens2024importance}, jeopardize participant safety \cite{PMID:30085607}, and lead to regulatory non-compliance \cite{ema2023seriousbreaches}. These reasons, among other factors for the failure of trials, result in an overall success rate of no more than 14\% for trials making it from phase I to the final market approval \cite{10.1093/biostatistics/kxx069}, prolonging the average 13.8-year-long drug development cycle \cite{martin2017trial}, as well as contributing to the average 1.3B\$ of drug development \cite{sertkaya2016key}.

In this study, we introduce an ML framework for early risk stratification of clinical trials (CTs) according to their likelihood of exhibiting a high rate of dosing errors, using information available prior to trial initiation. The proposed approach relies on data extracted from \href{https://clinicaltrials.gov}{ClinicalTrials.gov}, a large, publicly accessible registry of CTs maintained by the U.S. National Library of Medicine. This resource provides detailed protocol-level information, including eligibility criteria, and study design characteristics, alongside standardized adverse event reporting. Reporting to ClinicalTrials.gov is subject to statutory requirements for interventional drug trials, supporting the reliability and consistency of the reported data. This unique combination of scale, structure, and regulatory grounding makes this resource particularly well suited for large-scale, multimodal ML analyses of trial-level dosing error risk.

Our contributions are the following. First, we propose a multimodal ML framework for early, trial-level risk stratification of CTs according to their likelihood of exhibiting elevated dosing error rates. This approach integrates structured and semi-structured CT data and unstructured protocol-level text to support calibrated, trial-level stratification into actionable risk categories. Second, we publicly release the dataset used in this study, augmented with curated features and outcome labels, on \href{https://huggingface.co/datasets/ds4dh/ct-dosing-errors}{Hugging Face}. By making this resource openly available, we aim to facilitate reproducible research and stimulate further methodological development at the intersection of ML and CT safety. Finally, we provide the complete codebase used to construct the dataset and implement the proposed models on \href{https://github.com/ds4dh/CT-dosing-errors}{GitHub}. Particular emphasis was placed on reproducibility. We introduce a pipeline that enables automated ingestion of protocols with carefully designed data structures that capture the hierarchical nature of the CTGov's data fields, as well as strict validation of entries during parsing. This makes the reconstruction and extension of the dataset from the raw instances of ClinicalTrials.gov registry fully automated and reproducible.

This paper is organized as follows. Section~\ref{methodo} details the dataset construction process and the proposed ML methodology. Section~\ref{experiments} presents the experimental results. Section~\ref{discussion} interprets these findings and discusses their implications for CT safety and quality management. Section~\ref{conclusion} concludes the paper.

\section{Materials and Methods} \label{methodo}

\begin{figure}[h]
\centering
\includegraphics[width=16cm]{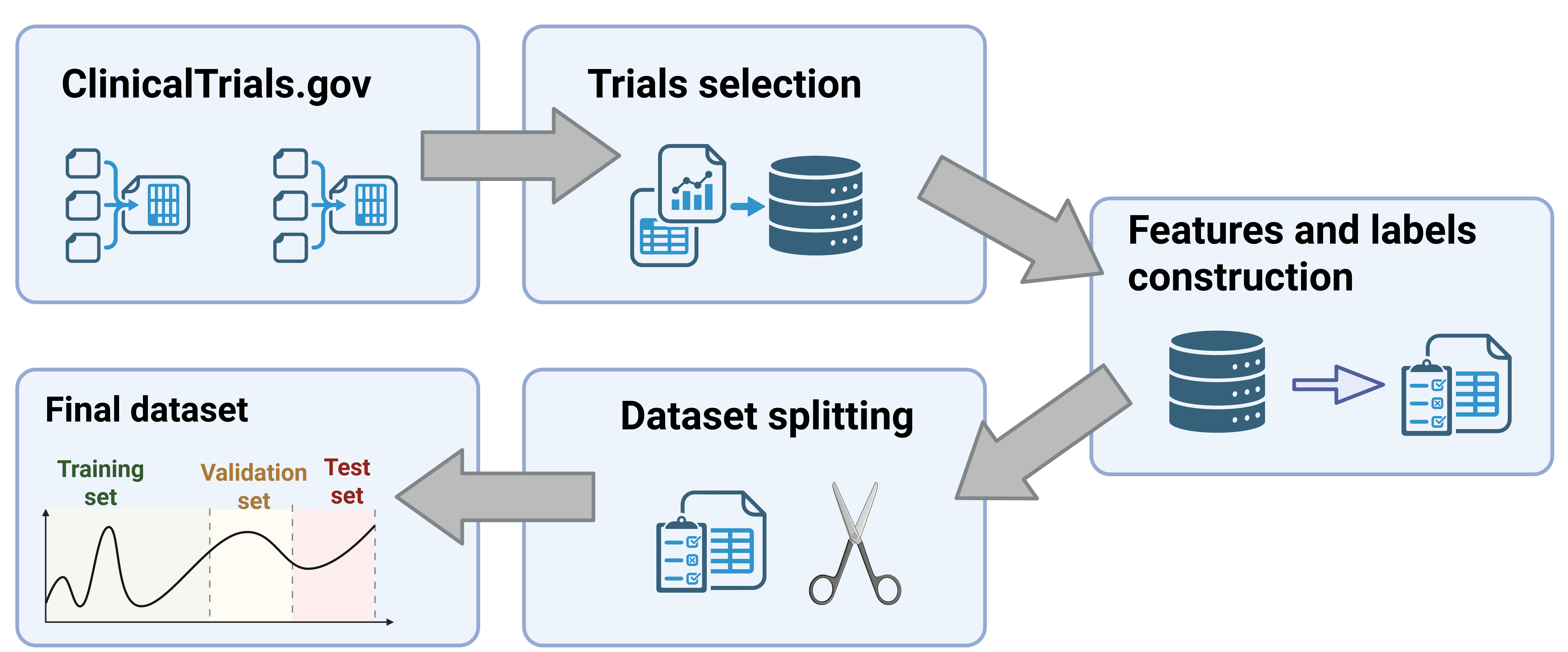}
\caption{Overview of the dataset construction steps, from CTs selection to feature and label construction and dataset splitting. Created with \href{https://BioRender.com}{BioRender}.}
\label{pipeline}
\end{figure}

Figure~\ref{pipeline} provides an overview of the dataset construction pipeline. Relevant CTs are first selected according to predefined inclusion criteria, as detailed in Section~\ref{trial selection}, notably to ensure the availability of dosing-related information. Information available prior to trial initiation is then extracted and ingested using customized data structures, from which predictive features are derived (Section~\ref{features}). The procedure used to assign dosing error–related risk labels is described in Section~\ref{labeling}.

The resulting dataset is subsequently partitioned into training, validation, and test sets using a chronological splitting strategy (Section~\ref{splitting strategy}). We then present the ML models and their evaluation procedures in Section~\ref{models}, followed by the trial-level risk assessment framework in Section~\ref{risk evaluation}.

\subsection{Trials selection} \label{trial selection}

The ClinicalTrials.gov registry was used as the primary source of clinical studies data, from which we construct our dataset. All CTs meeting the following inclusion criteria were retrieved: (i) completed or terminated trial status; (ii) interventional study design; (iii) availability of study results; (iv) a reported completion date; and (v) registration prior to a fixed cut-off date of September 1, 2025. In total, 42,112 CTs satisfied these criteria and were included in the dataset.

Restricting the dataset to completed or terminated trials guarantees that all included studies have concluded, while limiting the cohort to interventional designs focuses the analysis on therapeutic trials and excludes observational studies. Requiring available results ensures access to reported outcomes, such as adverse events. The availability of a completion date supports the temporal splitting strategy described in Section~\ref{splitting strategy}. Finally, the use of a fixed cut-off date ensures reproducibility.

\subsection{Data extraction and feature creation} \label{features}

All information fields available on ClinicalTrials.gov were screened to identify a relevant set of features. To reflect a realistic deployment scenario and avoid temporal leakage, only information available prior to trial initiation was retained as predictive input. To ensure data reliability and consistency, structured features were extracted using data validation tools such as the Pydantic library \cite{pydantic}, or Python's standard data types, such as the enumerated data structures (Enum) \cite{python_enum}, which constrain categorical variables to predefined value sets and prevent invalid or inconsistent entries. These schemas were created according to the structures defined in \href{https://clinicaltrials.gov/data-api/about-api/study-data-structure}{CTGov's instructions}.

The resulting feature set comprises categorical, numerical, and textual variables. An overview of these features is provided in Tables~\ref{tab:feature_overview}. Mappings between categorical encodings and their semantic meanings are provided in Appendix~\ref{features encoding}.

In addition to the predictive features and labels used in the present study, we extracted a set of auxiliary information to support flexible reuse of the dataset beyond its primary objective. These elements are intended to enable future investigations at the intersection of ML and CT safety, such as the construction of alternative datasets, analyses of trial design characteristics, or exploratory studies of discrepancies between enrollment and exposure populations. A description of this additional information is provided in Appendix~\ref{metada appendix}.

\begin{table*}[t]
\centering
\small
\setlength{\tabcolsep}{5pt}
\renewcommand{\arraystretch}{1.15}
\begin{tabular}{p{3.2cm}p{2.4cm}p{1.8cm}p{3.4cm}p{4.5cm}}
\toprule

\multicolumn{5}{l}{\textbf{Categorical and binary features}}\\[-2pt]
\midrule
\textbf{Feature name} &
\textbf{Data type} &
\textbf{Missing (\%)} &
\textbf{Most frequent class (\%)} &
\textbf{Description} \\
\midrule
primaryPurpose & Categorical & 0.94 & Treatment (84) & Primary objective of the study \\
masking & Categorical & 0.18 & None (52) & Masking strategy used in the trial design \\
sex & Categorical & 0.02 & All (88) & Sex eligibility criteria for participants \\
phases & Categorical (multi-label) & 0.01 & Phase II (41) & Clinical development phase(s) of the trial \\
armGroupTypes & Categorical (multi-label) & 0.60 & Experimental (64) & Type(s) of study arms \\
interventionTypes & Categorical (multi-label) & 0.00 & Drug (87) & Category(ies) of interventions evaluated \\
healthyVolunteers & Binary & 0.06 & False (86) & Eligibility of healthy volunteers for enrollment \\
oversightHasDmc & Binary & 12.00 & False (53) & Presence of a data monitoring committee \\

\addlinespace[3pt]
\midrule
\addlinespace[1pt]

\multicolumn{5}{l}{\textbf{Numeric features}}\\[-2pt]
\midrule
\textbf{Feature name} &
\textbf{Data type} &
\textbf{Missing (\%)} &
\textbf{Mean (SD)} &
\textbf{Description} \\
\midrule
enrollmentCount & Numeric (count) & 0.00 & 264 (4,810) & Number of enrolled participants \\
numArms & Numeric (count) & 0.00 & 2.33 (1.71) & Number of study arms per trial \\
numInterventions & Numeric (count) & 0.00 & 2.48 (1.68) & Number of distinct interventions evaluated \\
numLocations & Numeric (count) & 0.00 & 22.9 (62.9) & Number of trial sites or locations \\

\addlinespace[3pt]
\midrule
\addlinespace[1pt]

\multicolumn{5}{l}{\textbf{Text features}}\\[-2pt]
\midrule
\textbf{Feature name} &
\textbf{Data type} &
\textbf{Missing (\%)} &
\textbf{Mean characters (SD)} &
\textbf{Description} \\
\midrule
allocation & Free text & 0.38 & 8.46 (3.92) & Allocation method used to assign participants to study arms \\
interventionModel & Free text & 0.33 & 9.34 (1.78) & Intervention assignment model \\
briefSummary & Free text & 0.00 & 502.40 (464.74) & Short textual summary of the trial \\
detailedDescription & Free text & 38.00 & 1610.83 (1676.93) & Detailed description of objectives, design, and procedures \\
conditions & Free text & 0.00 & 39.21 (120.66) & Medical condition(s) studied \\
conditionsKeywords & Free text & 36.00 & 81.17 (155.22) & Additional condition-related keywords \\
armDescriptions & Free text & 0.60 & 388.24 (553.24) & Free-text descriptions of study arms \\
interventionNames & Free text & 0.00 & 82.72 (65.33) & Names of interventions reported in the trial \\
interventionDescriptions & Free text & 0.00 & 286.18 (340.14) & Descriptions of the interventions evaluated \\
locationDetails & Free text & 5.60 & 1310.26 (3512.07) & Textual information describing trial locations \\

\bottomrule
\end{tabular}
\vspace{1mm}
\caption{Overview of categorical/binary, numeric, and text features extracted from ClinicalTrials.gov. For each feature type, we report the corresponding summary statistic: most frequent class for categorical/binary variables, mean (SD) for numeric variables, and mean character length (SD) for text fields.}
\label{tab:feature_overview}
\end{table*}

\subsection{Label assignment} \label{labeling}

Dosing errors were identified using the standardized adverse event reporting framework of ClinicalTrials.gov, which relies on terminology from the Medical Dictionary for Regulatory Activities (MedDRA version 27.1, English) \cite{meddra}. MedDRA terms are organized hierarchically from general to more specific. Within this hierarchy, two High-Level Group Terms (HLGTs) related to dosing errors (\textit{10079145: ``Overdoses and underdoses NEC''} and \textit{10079159: ``Medication errors and other product use errors and issues''}) were selected. All the 630 descendants of these two nodes corresponding to the lower levels of the hierarchy, i.e., the High-Level Terms (HLTs), the Preferred Terms (PTs) and the Low-Level Terms (LLTs) were independently reviewed by a team of clinical pharmacology experts to focus only on the dosing-related events. This resulted in 210 concepts that we matched against the reported Adverse Events (AEs) of the included CT protocols using Fuzzy matching and canonicalized to 81 concepts.

For each trial, the reported CT's AEs were examined per Arm group, where we aggregated the number of participants at risk of experiencing adverse dosing events, as well as the actual occurring error numbers to the CT level and calculated dosing-error rates. Note that the number of participants may differ from the total number of enrolled participants, for example due to withdrawal prior to exposure. 

\begin{figure}[h]
\centering
\includegraphics[width=16cm]{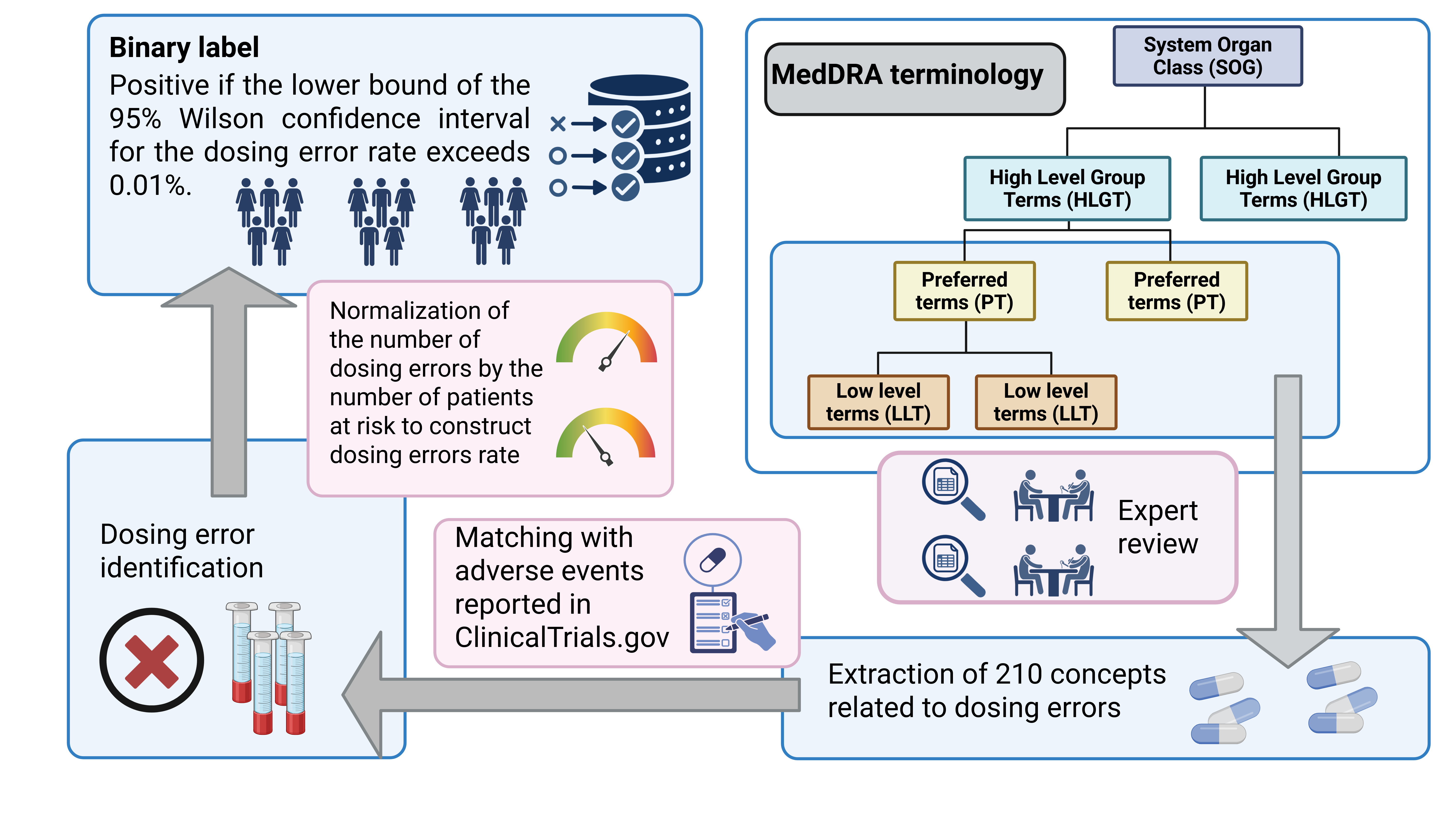}
\caption{Overview of the dosing-error labeling pipeline. MedDRA high-level group terms were selected and reviewed by clinical pharmacology experts, matched to reported adverse events, aggregated at the trial level, and converted into binary risk labels using a Wilson confidence interval–based threshold. Created with \href{https://BioRender.com}{BioRender}}
\label{medra organization}
\end{figure}

 Finally, a trial was labeled as positive if the lower bound of the 95\% Wilson confidence interval for the dosing error rate exceeded a fixed threshold. Given the rare nature of dosing errors in clinical research and the pharmacovigilance practice of treating very low-frequency events as meaningful safety signals \cite{beninger2020signal}, a low operational threshold of 0.01\% was adopted. Using this criterion, 4.62\% of the CTs in the dataset were labeled as positive, i.e., indicating an unusually high rate of dosing errors. The complete labeling workflow, including MedDRA term selection, expert curation, adverse event matching, and statistical thresholding, is summarized in Figure~\ref{medra organization}.

\subsection{Dataset splitting strategy} \label{splitting strategy}

A common practice when working with time-ordered data is to split observations chronologically and evaluate ML models on the most recent portion of the data, thereby providing a realistic estimate of prospective performance. Following this principle, we initially considered splitting the dataset based on trial initiation dates using a 70/15/15 train/validation/test split.

\begin{figure}[h]
\centering
\includegraphics[width=16cm]{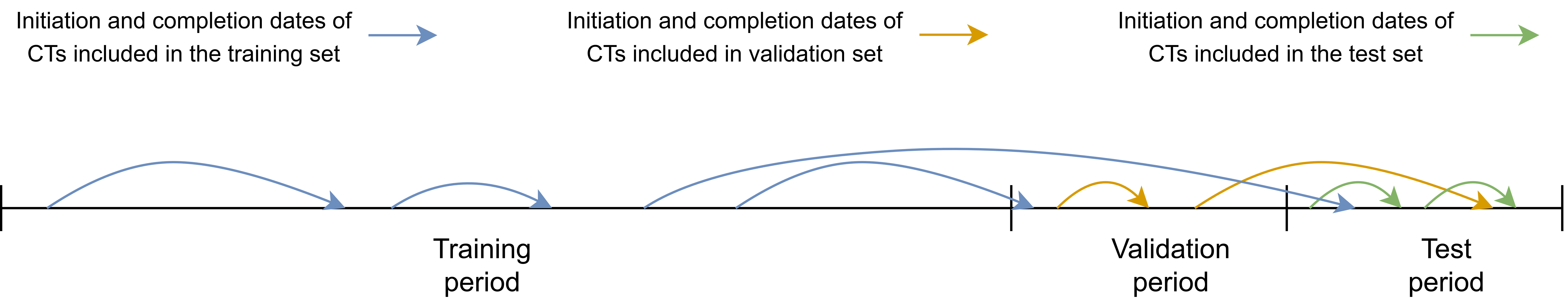}
\caption{Illustration of the bias introduced by splitting CTs chronologically according to initiation dates. Because only completed or terminated trials are included, trials with shorter durations are more likely to fall within later time windows, leading to their overrepresentation in the validation and test sets.}
\label{bias explanation}
\end{figure}

However, because the proposed pipeline relies exclusively on completed or terminated CTs, this strategy introduces a duration-dependent selection bias. As illustrated in Figure~\ref{bias explanation}, for trials initiated later in the study period, only those with sufficiently short durations are available in the dataset, as longer trials initiated during the same period have not yet completed or terminated. Consequently, trials with shorter durations, and by extension, smaller enrollment sizes, are overrepresented in the validation and test sets.

\begin{figure}[h]
\centering
\includegraphics[width=7cm]{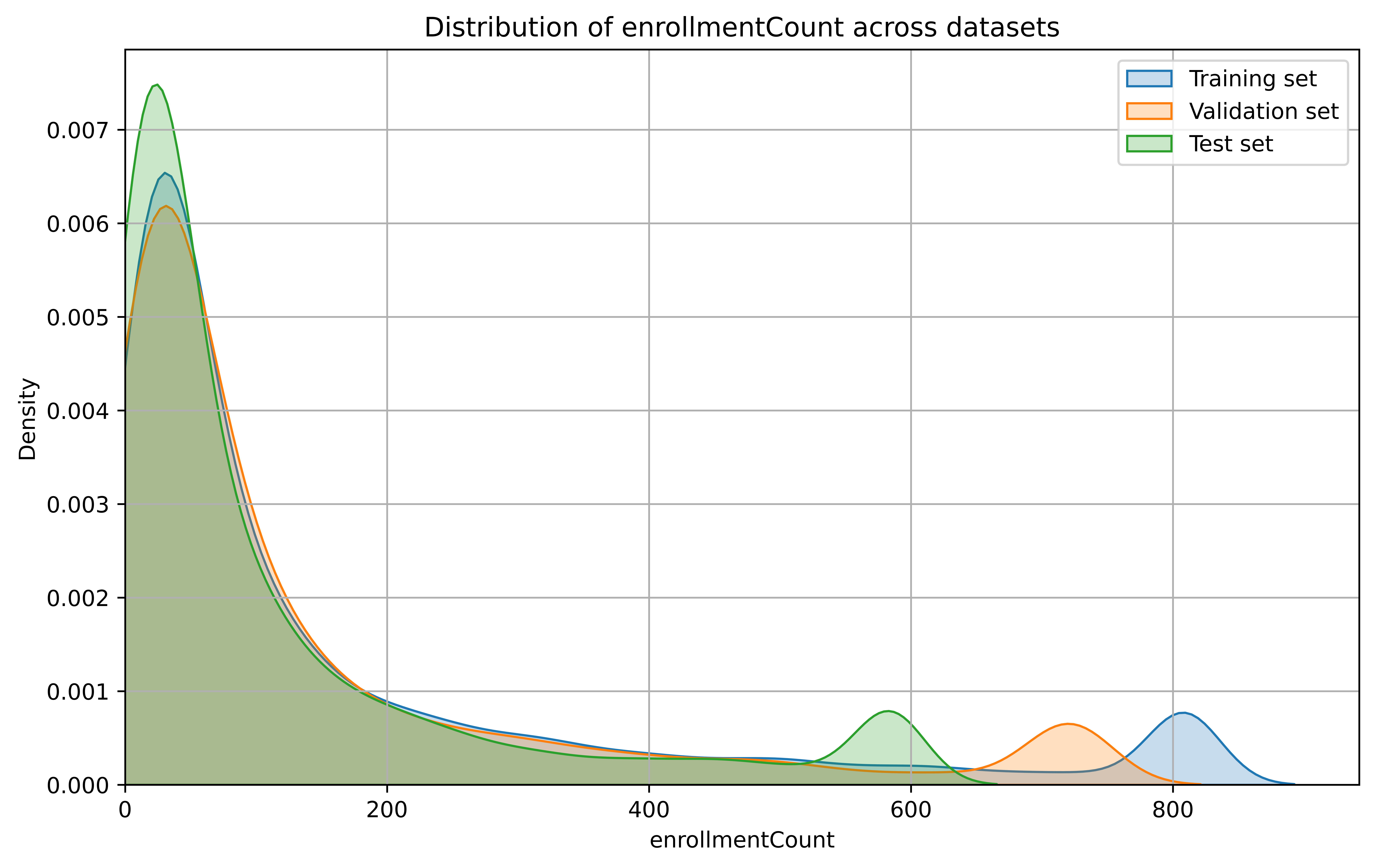}
\qquad
\includegraphics[width=7cm]{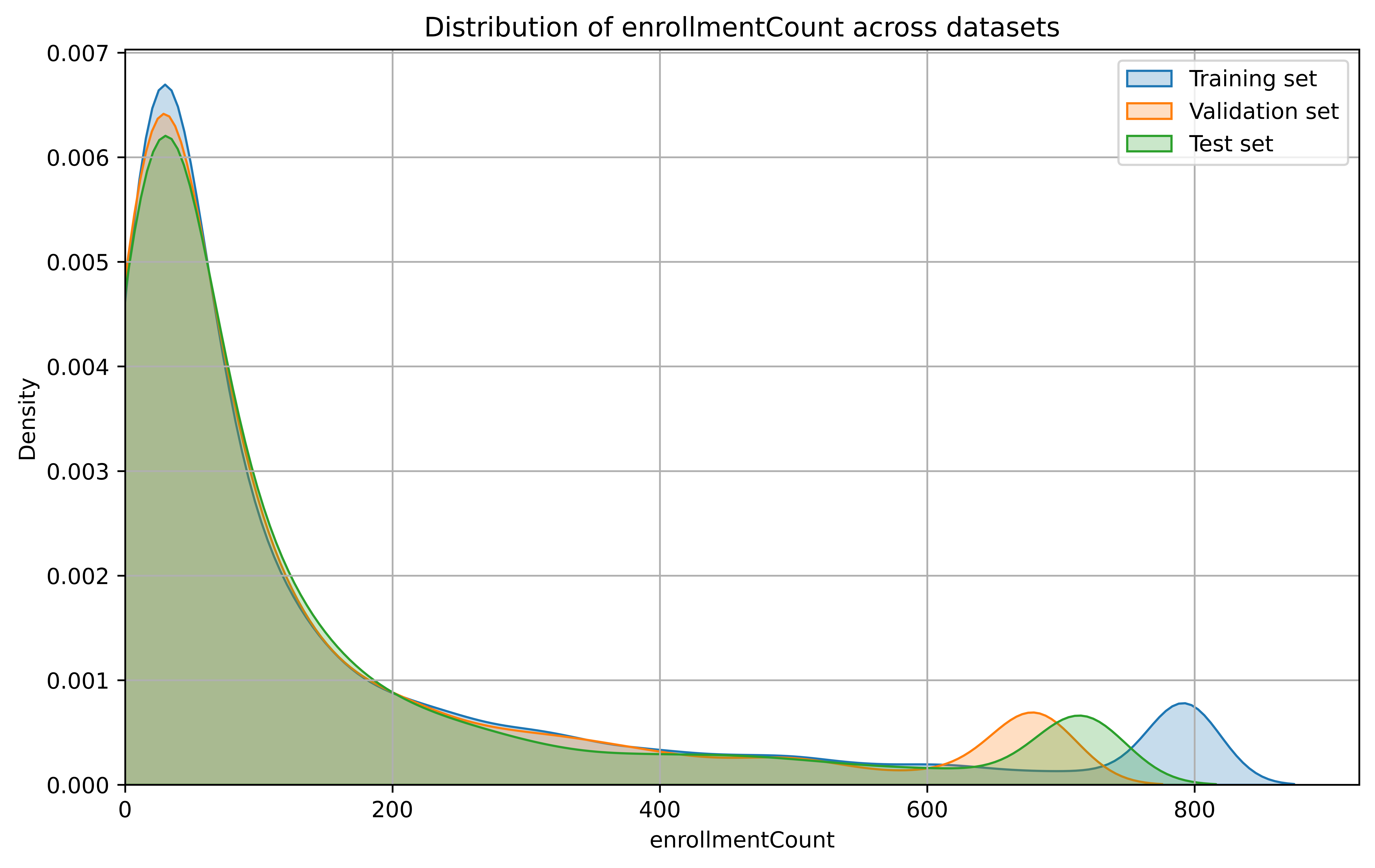}
\caption{Comparison of distributional shifts induced by different temporal splitting strategies. Kernel density estimates of enrollment counts across the training, validation, and test sets are shown for each strategy; distributions are truncated at the 95th percentile for visualization purposes. The left panel shows splitting by trial initiation date, and the right panel shows splitting by trial completion date, which we adopt in this work.}
\label{bias}
\end{figure}

To mitigate this bias, we instead sorted trials according to their completion dates prior to splitting. As shown in Figure~\ref{bias}, this approach reduces distributional shifts between the training, validation, and test sets. 

\subsection{Machine learning models} \label{models}

Our curated dataset contains fields with numerical, categorical, list-of-categoricals, and textual fields, where the latter may contain short textual descriptions with medical terminology. 
Our aim here is to establish a basic but robust baseline incorporating this multimodal information into the well-established predictive ML models.

Several standard ML models were trained and evaluated to assess their ability to identify CTs at risk of exhibiting a high dosing error rate. In addition to assessing predictive performance, these experiments were designed to examine whether different data modalities capture distinct and potentially complementary risk signals.

An XGBoost model \cite{chen2016xgboost} was trained using categorical and numerical features. A ClinicalModernBERT model \cite{lee2025clinical} was fine-tuned using only the text features. Finally, a simple multimodal approach, referred to as LateFusion model, was evaluated by combining the predicted probabilities of the two unimodal models using a weighted average, with weights optimized on the validation set. Details regarding training and hyperparameter optimization are provided in Appendix~\ref{implementation details}.

Model performance was primarily assessed using the area under the Receiver Operating Characteristic curve (AUC–ROC) to evaluate discrimination capacity. Additional metrics, including precision, recall, accuracy, and F1-score, were reported to provide a more comprehensive characterization of performance. Balanced accuracy and macro-averaged F1-score were computed to account for class imbalance. For metrics requiring a binary decision, the classification threshold was selected on the validation set to maximize the F1-score and then fixed for evaluation on the test set. Finally, the Brier score \cite{glenn1950verification} was used to assess the accuracy of predicted probabilities, providing a calibration-sensitive measure of probabilistic performance.

\subsection{Risk assessment} \label{risk evaluation}

In this study, we move beyond binary trial classification and focus on trial-level risk assessment for dosing errors. Rather than determining whether a CT will exhibit an elevated incidence of dosing errors, the objective is to provide a principled and interpretable framework for quantifying and stratifying risk across trials.

This shift from classification to risk estimation places strong requirements on the reliability of predicted probabilities. A naïve interpretation of ML model outputs as direct estimates of dosing error risk is inappropriate, as it is well established that ML models produce poorly calibrated probability estimates \cite{guo2017calibration, van2019calibration}. Raw model outputs cannot be reliably interpreted as absolute risk estimates and may substantially misrepresent true risk levels.

To address this limitation, we applied post-hoc probability calibration prior to downstream risk assessment. Calibrated probabilities were obtained using Platt scaling \cite{platt1999probabilistic} for the ClinicalModernBERT and LateFusion models, and isotonic regression \cite{zadrozny2002transforming} for the XGBoost model. Calibration parameters were estimated exclusively on the validation set and subsequently applied unchanged to the test set.

The calibrated predicted probabilities were then used to stratify CTs into discrete risk categories reflecting their likelihood of exhibiting a high incidence of dosing errors. In the full dataset, approximately 4.6\% of trials were classified as risky, providing an empirical estimate of the average baseline risk. Risk thresholds were therefore defined a priori to correspond to increasing multiples of this baseline, enabling a direct and intuitive interpretation of risk levels.

Specifically, the categories were constructed based on the predicted probability $\hat{p}$ as follows:
\begin{itemize}
    \item \textbf{Low risk}: $\hat{p} < 2\%$ (less than $0.5\times$ the average risk),
    \item \textbf{Moderate risk}: $2\% \leq \hat{p} < 5\%$ (approximately $0.5$--$1\times$ the average risk),
    \item \textbf{High risk}: $5\% \leq \hat{p} < 10\%$ (approximately $1$--$2\times$ the average risk),
    \item \textbf{Very high risk}: $\hat{p} \geq 10\%$ (greater than $2\times$ the average risk).
\end{itemize}

To evaluate the practical utility of the proposed risk categories, we report, for each model and each risk group, the number of assigned trials, the number of trials classified as problematic, the observed event rate, and the corresponding risk increase relative to the overall rate observed in the test set.

To further assess the robustness of the proposed risk stratification framework, we performed subgroup analyses across trial characteristics. Specifically, predicted risk groups were examined according to clinical development stage and enrollment size. Development stage was determined from the highest reported trial phase and mapped into three categories (early: Early Phase I or Phase I; mid: Phase II; late: Phase III or beyond). Enrollment size was grouped into predefined bins ($\leq$50, 51–200, 201–500, and >500 participants). Within each subgroup and predicted risk category, we computed the number of trials, the number of problematic trials, the observed event rate, and the relative risk relative to the subgroup-specific baseline prevalence.

\section{Results} \label{experiments}

Table~\ref{tab:classification_results} reports test-set classification performance for the evaluated ML models, including both uncalibrated and calibrated variants. Among the unimodal approaches, XGBoost obtained an AUC-ROC of 0.848, while ClinicalModernBERT achieved an AUC-ROC of 0.855. The LateFusion model yielded the highest AUC-ROC (0.862).

Across all evaluated models, post-hoc probability calibration substantially improved probabilistic accuracy, reducing Brier scores from approximately 0.09–0.11 to 0.04–0.05. As expected, calibration did not materially affect discriminative performance, with AUC-ROC values remaining unchanged.

\begin{table}[t]
\centering
\renewcommand{\arraystretch}{1.05}
\begin{tabular*}{\textwidth}{@{\extracolsep{\fill}}lcccccccc@{}}
\toprule
\textbf{Model} 
& \textbf{AUC} 
& \textbf{Brier} 
& \textbf{F1} 
& \textbf{F1 (macro)} 
& \textbf{Rec} 
& \textbf{Prec} 
& \textbf{BalAcc} 
& \textbf{Acc} \\
\midrule

XGBoost (uncal.)
& 0.848 & 0.105 & 0.293 & 0.619 & 0.423 & 0.224 & 0.674 & 0.900 \\

XGBoost
& 0.848 & 0.042 & 0.292 & 0.619 & 0.423 & 0.222 & 0.674 & 0.899 \\

\midrule

ClinicalModernBERT (uncal.)
& 0.855 & 0.092 & 0.304 & 0.626 & \textbf{0.429} & 0.235 & \textbf{0.678} & 0.903 \\

ClinicalModernBERT
& 0.855 & 0.042 & \textbf{0.306} & 0.628 & 0.426 & 0.239 & \textbf{0.678} & 0.905 \\

\midrule

LateFusion (uncal.)
& \textbf{0.862} & 0.090 & 0.300 & \textbf{0.633} & 0.284 & 0.318 & 0.626 & \textbf{0.935} \\

LateFusion
& \textbf{0.862} & \textbf{0.041} & 0.298 & 0.632 & 0.281 & \textbf{0.319} & 0.625 & \textbf{0.935} \\

\bottomrule
\end{tabular*}
\vspace{1mm}
\caption{Classification performance on the test set before and after probability calibration. Best performance per metric is shown in bold.}
\label{tab:classification_results}
\end{table}

\begin{table}[htbp]
\centering
\small
\begin{tabular*}{\textwidth}{@{\extracolsep{\fill}}lcccc}
\toprule
Risk group 
& Number of CTs 
& Number of events 
& Event rate (\%) 
& Relative risk \\
\midrule
\multicolumn{5}{l}{\textbf{LateFusion (uncal.)}} \\
Low        & 401  & 0 & 0.00  & 0.000  \\
Moderate   & 1773 & 6 & 0.34  & 0.069   \\
High       & 953 & 8 & 0.84 & 0.171 \\
Very high  & 3191 & 296 & 9.28 & 1.89 \\
\midrule
\addlinespace
\multicolumn{5}{l}{\textbf{LateFusion}} \\
Low        & 3547 & 22  & 0.62  & 0.126  \\
Moderate   & 948 & 26 & 2.74  & 0.559  \\
High       & 738 & 58 & 7.86 &  1.602 \\
Very high  & 1085 & 204 & 18.80 & 3.832  \\
\bottomrule
\end{tabular*}
\vspace{1mm}
\caption{
Risk stratification of CTs on the test set using calibrated LateFusion predictions. Events correspond to trials labeled as exhibiting an elevated dosing error rate. Relative risks are reported relative to the overall event prevalence in the test set (4.91\%).
}
\label{tab:risk_stratification_models}
\end{table}

Given that the calibrated LateFusion model achieved both the highest discriminative performance and the lowest Brier score among the evaluated approaches, we focus on this model for downstream risk stratification. Table~\ref{tab:risk_stratification_models} summarizes the distribution of CTs across predicted risk groups on the test set, together with observed event rates and relative risks. Results for the remaining models are reported in the Appendix ~\ref{all risks}.

Notably, the results obtained with the calibrated LateFusion model demonstrate a clear alignment between predefined risk categories and observed outcomes. Among CTs classified as low risk ($\hat{p} < 2\%$), 0.62\% were labeled as problematic. This proportion increases to 2.74\% for trials in the moderate-risk group ($2\% \leq \hat{p} < 5\%$), 7.86\% in the high-risk group ($5\% \leq \hat{p} < 10\%$), and 18.80\% in the very high–risk group ($\hat{p} \geq 10\%$). In addition, the proposed stratification distributes CTs across all predefined risk categories, avoiding collapse into a single dominant class and yielding non-degenerate risk groups. In contrast, risk stratification based on uncalibrated model outputs resulted in a markedly less informative allocation of trials across risk groups.

Subgroup analyses across development stage and enrollment size revealed consistent and informative patterns. As shown in Table~\ref{tab:risk_stratification_by_stage}, the monotonic increase in observed event rates across risk categories derived from the calibrated LateFusion model was preserved within early-, mid-, and late-stage trials. Within each development stage, event rates increased stepwise from the low- to the very high–risk groups defined by the model’s predicted probabilities, and the corresponding relative risks exhibited comparable monotonic increases. These findings indicate that risk categories maintain their discriminatory structure independently of clinical phase. Complementary analyses stratified by enrollment size (Appendix~\ref{risk by enrollment}) demonstrated a similar progressive enrichment of problematic trials across higher risk groups.

\section{Discussion} \label{discussion}

These results suggest that dosing error risk in CTs can be anticipated using information that is available prior to trial initiation. By relying exclusively on CT design characteristics and protocol-related textual information, the proposed framework indicates that aspects of dosing error risk are associated with upstream design and planning decisions, supporting a preventive perspective on medication safety in clinical research.

Across models, multimodal integration of structured trial features and unstructured protocol text outperformed unimodal baselines. Notably, these improvements were achieved using a simple late-fusion strategy, without relying on complex multimodal architectures. This suggests that structured trial information and free-text protocol descriptions capture distinct and complementary aspects of dosing error risk, and that their integration can improve trial-level risk estimation.

\begin{table}[htbp]
\centering
\small
\begin{tabular*}{\textwidth}{@{\extracolsep{\fill}}lcccc}
\toprule
Risk group 
& Number of CTs 
& Number of events 
& Event rate (\%) 
& Relative risk \\
\midrule

\multicolumn{5}{l}{\textbf{Early stage}} \\
Low        & 642  & 3   & 0.47  & 0.235 \\
Moderate   & 75   & 1   & 1.33  & 0.672 \\
High       & 50   & 5   & 10.00 & 5.038 \\
Very high  & 39   & 7   & 17.95 & 9.042 \\
\midrule
\addlinespace

\multicolumn{5}{l}{\textbf{Mid stage}} \\
Low        & 1853 & 14  & 0.76  & 0.201 \\
Moderate   & 526  & 15  & 2.85  & 0.758 \\
High       & 352  & 32  & 9.09  & 2.417 \\
Very high  & 326  & 54  & 16.56 & 4.403 \\
\midrule
\addlinespace

\multicolumn{5}{l}{\textbf{Late stage}} \\
Low        & 934  & 5   & 0.54  & 0.070 \\
Moderate   & 344  & 10  & 2.91  & 0.379 \\
High       & 335  & 21  & 6.27  & 0.817 \\
Very high  & 719  & 143 & 19.89 & 2.591 \\
\bottomrule
\end{tabular*}
\vspace{1mm}
\caption{
Risk stratification by clinical development stage on the test set using calibrated LateFusion predictions. Relative risks are reported with respect to the prevalence within each development stage.
}
\label{tab:risk_stratification_by_stage}
\end{table}

Beyond discriminative performance, a key methodological insight concerns the role of probability calibration in enabling reliable and interpretable risk stratification. Models incorporating calibrated outputs yielded a partition of CTs into distinct risk categories, avoiding collapse into a single dominant class and exhibiting a clear, monotonic relationship between predicted risk levels and observed CTs with unusually high dosing error rates. In contrast, stratification based on uncalibrated model outputs resulted in substantially less informative groupings.

In addition, the subgroup analyses further support the structural robustness of the proposed framework. The preservation of monotonic risk trends within development phases and across enrollment categories suggests that the observed stratification cannot be reduced to simple proxies such as clinical stage or trial size. Rather, the model appears to capture more granular design-level characteristics that contribute to dosing error risk beyond these structural attributes.

\subsection{Implications for clinical trial quality management}

The ability to anticipate dosing error risk using information available prior to trial initiation has direct implications for CT quality management. By providing an interpretable, trial-level risk signal early in the study lifecycle, the proposed framework enables a more proactive and proportionate approach to risk mitigation. Notably, the preservation of risk enrichment patterns across development phases and enrollment sizes suggests that the approach remains informative across heterogeneous clinical contexts, including trials at different stages of development and of varying scale. Rather than applying uniform review processes across all trials, risk stratification therefore allows preventive efforts to be tailored according to the estimated likelihood of dosing-related issues, independent of trial phase or size.

The proposed framework could be integrated into existing CT review and governance workflows as a decision-support tool during the planning phase. Predicted risk categories provide a quantitative and interpretable signal that can complement established review criteria. In practice, this information may inform targeted protocol refinements, particularly with respect to dosing schedules, administration procedures, or monitoring plans, or motivate the implementation of additional safeguards prior to trial initiation. Importantly, the framework is not intended to replace expert judgment, but rather to augment it by providing a data-driven risk signal that is available early in the trial lifecycle.

More broadly, these findings suggest that aspects of dosing error risk are encoded in upstream trial design and protocol characteristics, reinforcing the value of preventive, design-stage interventions for improving medication safety in clinical research. By enabling early identification of trials that warrant closer scrutiny, the proposed approach supports a proactive shift from reactive error detection toward anticipatory risk management.

\subsection{Limitations and future directions}

This study has a few limitations that warrant consideration. First, dosing error labels were derived from adverse event reports in ClinicalTrials.gov and may therefore be affected by underreporting. Nevertheless, the use of a large-scale dataset drawn from a single, standardized reporting system provides a consistent basis for comparative analysis across a heterogeneous population of CTs, supporting the identification of trial-level risk patterns within the scope of the available data.

Second, although a multimodal modeling strategy was adopted, the integration of structured and unstructured information was implemented using a deliberately simple late-fusion approach. More sophisticated multimodal architectures may warrant further investigation and could potentially yield additional performance gains. However, the improvements observed with this basic fusion strategy suggest that the different data modalities capture complementary risk-related information, indicating that multimodal integration remains beneficial even under conservative modeling assumptions.

Third, while full protocol documents were collected and converted into machine-readable text, the present study did
not exploit this source of information.
Such documents often contain detailed dosing procedures and other relevant clinical details but pose substantial methodological challenges due to their length, heterogeneity, and complex structure. Although addressing these challenges represents an exciting direction for future research, it was beyond the scope of the present study. In this context, the dataset and processing pipeline introduced here provide a foundation for further investigation. The dataset has additionally been released as part of the \href{https://www.codabench.org/competitions/11891/}{CT-DEB 2026 shared task}, hosted within the CL4Health workshop at LREC 2026, to stimulate further research on protocol-level document analysis for CT safety.

\section{Conclusion} \label{conclusion}

In this study, we introduced an ML framework for early, trial-level stratification of dosing error risk in CTs. By enabling straightforward adaptation to other categories of medication errors, alternative definitions of acceptable error rate thresholds, and different CT inclusion criteria, the proposed framework can be readily tailored to specific regulatory, operational, or safety objectives. The approach leverages information available prior to trial initiation to support proactive risk assessment at the planning stage. Experimental results demonstrate that the framework can stratify CTs according to dosing error risk using pre-initiation information. Importantly, the observed frequency of trials exhibiting high dosing error rates within each predicted risk group was aligned with the predefined probability bounds used to define these categories, supporting the reliability of the resulting risk stratification. Moreover, the preservation of risk trends across development phases and enrollment categories further supports the structural robustness of the proposed stratification framework.

This study highlights two key methodological insights. First, multimodal integration of structured trial information and unstructured protocol text consistently improved trial-level discrimination performance, even when implemented using a deliberately simple late-fusion strategy, underscoring the complementary nature of these data sources. Second, probability calibration appeared to be critical for aligning predicted risks with interpretable probability thresholds, enabling reliable and flexible risk stratification beyond binary classification. From a practical perspective, the proposed framework supports early, trial-level risk assessment that may inform proactive quality management in clinical research. By providing interpretable risk categories prior to trial initiation, the approach may support the identification of trials that warrant additional review, protocol refinement, or targeted quality safeguards.

A key strength of the proposed framework lies in its reproducibility, scalability, and interpretability. All components of the pipeline, from data extraction and label construction to model training and evaluation, are publicly released and rely exclusively on publicly available data sources, enabling straightforward regeneration of the dataset under alternative inclusion criteria or temporal cut-offs and facilitating reuse across a broad range of research activities related to CT safety. This infrastructure provides a foundation for future methodological extensions, including the exploration of more sophisticated multimodal architectures and the exploitation of full protocol documents. Ultimately, early, data-driven trial-level risk assessment represents a promising direction for supporting improvements in the safety and quality of clinical research.

\section*{Author contributions}

\textbf{Félicien Hêche:} Conceptualization; Methodology;
 Investigation; Formal analysis; Writing -- original draft; 
\textbf{Sohrab Ferdowsi:} Conceptualization; Methodology; Data curation; Investigation; 
Writing -- review \& editing.
\textbf{Anthony Yazdani:} Methodology; Data curation; 
Writing -- review \& editing.
\textbf{Sara Sansaloni-Pastor:} 
Writing -- review \& editing.
\textbf{Douglas Teodoro:} Conceptualization; Supervision; 
Funding acquisition; Writing -- review \& editing.

\section*{Funding statement}
This work was supported by Innosuisse – the Swiss Innovation Agency – grant number 114.721 IP-ICT

\section*{Conflict of interest}
Sara Sansaloni-Pastor works for Actelion Pharmaceuticals Ltd. The other authors declare no competing interests.

\section*{Data availability}

All source data were obtained from the publicly accessible ClinicalTrials.gov registry. 
The curated dataset generated in this study is publicly available on 
\href{https://huggingface.co/datasets/ds4dh/ct-dosing-errors}{Hugging Face}. 
The full analysis pipeline and model implementation are available on 
\href{https://github.com/ds4dh/CT-dosing-errors}{GitHub}.

\bibliographystyle{unsrt}
\bibliography{references}

\appendix

\section{Categorical features encoding} \label{features encoding}

This appendix describes the encoding of some of the categorical structured features extracted from ClinicalTrials.gov. Each feature was mapped to predefined integer identifiers representing distinct categories, ensuring consistent and interpretable model inputs. The resulting mappings are reported in Table~\ref{tab:cat_mappings}.


\begin{table}[h!]
\centering
\footnotesize
\begin{tabular}{p{3cm}p{1.2cm}p{5.5cm}}
\hline
\textbf{Feature} & \textbf{ID} & \textbf{Meaning} \\
\hline
phases & 0 & NA (not reported) \\
      & 1 & EARLY\_PHASE1 \\
      & 2 & PHASE1 \\
      & 3 & PHASE2 \\
      & 4 & PHASE3 \\
      & 5 & PHASE4 \\

\hline
primaryPurpose & 0 & TREATMENT \\
              & 1 & PREVENTION \\
              & 2 & DIAGNOSTIC \\
              & 3 & ECT \\
              & 4 & SUPPORTIVE\_CARE \\
              & 5 & SCREENING \\
              & 6 & HEALTH\_SERVICES\_RESEARCH \\
              & 7 & BASIC\_SCIENCE \\
              & 8 & DEVICE\_FEASIBILITY \\
              & 9 & OTHER \\

\hline
masking & 0 & NONE \\
        & 1 & SINGLE \\
        & 2 & DOUBLE \\
        & 3 & TRIPLE \\
        & 4 & QUADRUPLE \\

\hline
sex & 0 & ALL \\
    & 1 & FEMALE \\
    & 2 & MALE \\

\hline
armGroupTypes & 0 & EXPERIMENTAL \\
             & 1 & ACTIVE\_COMPARATOR \\
             & 2 & PLACEBO\_COMPARATOR \\
             & 3 & SHAM\_COMPARATOR \\
             & 4 & NO\_INTERVENTION \\
             & 5 & OTHER \\

\hline
interventionTypes & 0 & DRUG \\
                  & 1 & DEVICE \\
                  & 2 & BIOLOGICAL \\
                  & 3 & PROCEDURE \\
                  & 4 & RADIATION \\
                  & 5 & BEHAVIORAL \\
                  & 6 & GENETIC \\
                  & 7 & DIETARY\_SUPPLEMENT \\
                  & 8 & COMBINATION\_PRODUCT \\
                  & 9 & DIAGNOSTIC\_TEST \\
                  & 10 & OTHER \\
\hline
\end{tabular}
\vspace{1mm}
\caption{Overview of mappings between categorical features and their corresponding integer encodings. For multi-label variables, multiple category identifiers may be associated with a single trial.}
\label{tab:cat_mappings}
\end{table}

\FloatBarrier 

\section{Additional extracted information} \label{metada appendix}

This appendix describes the auxiliary information extracted in parallel with the features and labels used in the present study. These variables were not incorporated into the primary modeling framework but were curated to facilitate regeneration of the dataset under alternative inclusion criteria or labeling strategies, and enable secondary analyses related to CT safety and design characteristics. A complete overview of these elements is provided in Table~\ref{tab:metadata_overview}. In addition to fields directly available from ClinicalTrials.gov, full trial protocols, typically provided as lengthy PDF documents, were also downloaded and extracted from the links provided within the trials. These downloaded PDF documents were then converted into machine-readable text using PyMuPDF~\cite{pymupdf}. The accompanying codebase further supports optical character recognition–based extraction via DeepSeek-OCR~\cite{wei2025deepseek} for more careful serialization.

\begin{table}[h!]
\centering
\small
\begin{tabular}{p{4.0cm}p{3.2cm}p{6.8cm}}
\hline
\textbf{Name} & \textbf{Data type / encoding} & \textbf{Description} \\
\hline

overallStatus & Categorical &
Overall recruitment status of the trial \\

leadSponsorClass & Categorical &
Sponsor class \\

hasProtocol & Binary indicator &
Indicates whether a protocol document is available \\

hasSap & Binary indicator &
Indicates whether a statistical analysis plan document is available \\

hasIcf & Binary indicator &
Indicates whether an informed consent form document is available \\

startDate & Date &
Trial initiation date \\

completionDate & Date &
Trial completion date \\

nctId & String &
ClinicalTrials.gov unique trial identifier (NCT number) \\

leadSponsorName & String &
Name of the lead sponsor organization \\

protocolPdfLinks & String  &
Links to protocol PDF document(s) when available \\

protocolPdfText & String & Machine-readable text extracted from available trial protocol PDF document(s) \\

count\_\textit{X} & Numeric (count) &
Family of variables capturing the number of adverse event reports mapped to MedDRA dosing-error term \textit{X} \\

wilson\_lower\_bound & Numeric (continuous) &
Lower bound of the 95\% Wilson confidence interval for the dosing error rate \\

ct\_level\_ade\_population & Numeric (count) &
Number of participants at risk of experiencing an adverse drug event \\

sum\_dosing\_errors & Numeric (count) &
Number of dosing errors \\

dosing\_error\_rate & Numeric (continuous) &
Number of dosing errors normalized by the number of patients at risk \\

\hline
\end{tabular}
\vspace{1mm}
\caption{Overview of additional information extracted.}
\label{tab:metadata_overview}
\end{table}

\FloatBarrier 

\section{Implementation details} \label{implementation details}

Hyperparameters of the XGBoost model were optimized using Optuna~\cite{akiba2019optuna}, with the search space summarized in Table~\ref{tab:xgb_hyperparameter_space}. A total of 200 hyperparameter configurations were evaluated, and the model achieving the best AUC-ROC score on the validation set was selected for final evaluation on the test set. To address class imbalance, positive samples were upweighted via a class-weighting parameter, which was jointly optimized during hyperparameter tuning. Missing values in structured features were managed using XGBoost’s native missing-value mechanism.

The ClinicalModernBERT model was trained with a maximum context length of 8,192 tokens, a learning rate of $2.5 \times 10^{-5}$, a weight decay of $0.01$, and an effective batch size of 32 achieved through gradient accumulation over four steps. Hyperparameters were not tuned but selected based on prior experience with this architecture. Training was conducted for up to 100 epochs, with early stopping applied using a patience of five epochs. To mitigate class imbalance, class-balanced mini-batch sampling was employed, enforcing a fixed proportion of $0.5$ positive samples per batch. For text fields with missing values, empty entries were replaced with a dedicated string (“UNKNOWN”) prior to tokenization.

For the LateFusion model, the fusion weight used to combine the unimodal predicted probabilities was selected to maximize the AUC–ROC score on the validation set using Optuna over 100 trials.

\begin{table}[h!]
\centering
\small
\begin{tabular}{p{4.2cm}p{4.0cm}p{5.2cm}}
\hline
\textbf{Hyperparameter} & \textbf{Search range} & \textbf{Description} \\
\hline

n\_estimators & 100 -- 1000 &
Number of boosting trees \\

max\_depth & 3 -- 12 &
Maximum depth of individual decision trees \\

learning\_rate & $[10^{-3}, 0.3]$ &
Learning rate controlling the contribution of each tree; sampled on a logarithmic scale \\

subsample & 0.5 -- 1.0 &
Fraction of training samples randomly subsampled for each tree \\

colsample\_bytree & 0.5 -- 1.0 &
Fraction of features randomly subsampled for each tree \\

gamma & 0 -- 5 &
Minimum loss reduction required to make a further partition on a leaf node \\

min\_child\_weight & 1 -- 10 &
Minimum sum of instance weights required in a child node \\

max\_delta\_step & 0 -- 10 &
Maximum change in leaf weights allowed per boosting iteration \\

reg\_alpha & $[10^{-8}, 10]$ &
L1 regularization strength on model weights; sampled on a logarithmic scale  \\

reg\_lambda & $[10^{-8}, 10]$ &
L2 regularization strength on model weights; sampled on a logarithmic scale \\

scale\_pos\_weight & $[0.5w, 2.0w]$ &
Class imbalance weighting, where $w$ denotes the empirical ratio of negative to positive samples \\

\hline
\end{tabular}
\vspace{1mm}
\caption{XGBoost hyperparameter search space.}
\label{tab:xgb_hyperparameter_space}
\end{table}

\FloatBarrier

\section{Additional results} \label{all risks}

This appendix reports the risk stratification results for all evaluated ML models on the test set. Table~\ref{tab:all_risk_stratification} summarizes the distribution of CTs across predefined risk groups for both calibrated and uncalibrated model variants.

\begin{table}[htbp]
\centering
\small
\begin{tabular*}{\textwidth}{@{\extracolsep{\fill}}lcccc}
\toprule
& Number of CTs 
& Number of events 
& Event rate (\%) 
& Relative risk \\
\midrule
\multicolumn{5}{l}{\textbf{XGBoost (uncal.)}} \\
Low        & 79 &  0  & 0.00  & 0.000  \\
Moderate   & 1311 & 1 & 0.08 &  0.015 \\
High       & 1252 & 8 & 0.64 & 0.130 \\
Very high  & 3676 & 301 & 8.19 & 1.669 \\
\midrule
\multicolumn{5}{l}{\textbf{XGBoost}} \\
Low        & 3422 & 22  & 0.64   & 0.131 \\
Moderate   & 896 & 24 & 2.68 & 0.546 \\
High       & 975 & 79  & 8.10  & 1.651 \\
Very high  & 1025 & 185 & 18.05 & 3.68 \\
\midrule
\multicolumn{5}{l}{\textbf{ClinicalModernBERT (uncal.)}} \\
Low        & 4002  & 33  & 0.83   & 0.168 \\
Moderate   & 391 & 17 & 4.35 & 0.886 \\
High       & 253 & 10 & 3.95 & 0.806 \\
Very high  & 1672 & 250 & 14.95 & 3.047 \\
\midrule
\multicolumn{5}{l}{\textbf{ClinicalModernBERT}} \\
Low        & 3741 & 24 & 0.64  & 0.131 \\
Moderate   & 853 & 33 & 3.87 & 0.788 \\
High       & 520 & 41 & 7.88 & 1.607\\
Very high  & 1204 & 212 & 17.61 & 3.589 \\
\bottomrule
\end{tabular*}
\vspace{1mm}
\caption{
Risk stratification results for all evaluated models on the test set. Results are shown for both calibrated and uncalibrated variants using predefined risk group cut-offs applied consistently across models.
}
\label{tab:all_risk_stratification}
\end{table}

\section{Risk stratification by enrollment size} \label{risk by enrollment}

This appendix reports the results of the risk stratification analysis according to trial enrollment size. Table~\ref{tab:risk_stratification_by_enrollment} summarizes the distribution of CTs across predefined risk categories within each enrollment bin ($\leq$50, 51–200, 201–500, and >500 participants) on the test set. For each category, we report the number of trials, the number of problematic trials, the observed event rate, and the relative risk computed with respect to the baseline prevalence within the corresponding enrollment group.

\begin{table}[htbp]
\centering
\small
\begin{tabular*}{\textwidth}{@{\extracolsep{\fill}}lcccc}
\toprule
Risk group 
& Number of CTs 
& Number of events 
& Event rate (\%) 
& Relative risk \\
\midrule

\multicolumn{5}{l}{\textbf{Enrollment $\leq$ 50}} \\
Low        & 2535 & 16 & 0.63  & 0.517 \\
Moderate   & 328  & 5  & 1.52  & 1.248 \\
High       & 124  & 8  & 6.45  & 5.283 \\
Very high  & 43   & 8  & 18.60 & 15.236 \\
\midrule
\addlinespace

\multicolumn{5}{l}{\textbf{Enrollment 51--200}} \\
Low        & 912  & 4  & 0.44  & 0.099 \\
Moderate   & 421  & 13 & 3.09  & 0.698 \\
High       & 354  & 32 & 9.04  & 2.042 \\
Very high  & 301  & 39 & 12.96 & 2.927 \\
\midrule
\addlinespace

\multicolumn{5}{l}{\textbf{Enrollment 201--500}} \\
Low        & 92   & 2  & 2.17  & 0.198 \\
Moderate   & 149  & 8  & 5.37  & 0.488 \\
High       & 191  & 13 & 6.81  & 0.619 \\
Very high  & 332  & 61 & 18.37 & 1.671 \\
\midrule
\addlinespace

\multicolumn{5}{l}{\textbf{Enrollment $>$ 500}} \\
Low        & 8    & 0  & 0.00  & 0.000 \\
Moderate   & 50   & 0  & 0.00  & 0.000 \\
High       & 69   & 5  & 7.25  & 0.385 \\
Very high  & 409  & 96 & 23.47 & 1.246 \\
\bottomrule
\end{tabular*}
\vspace{1mm}
\caption{
Risk stratification by enrollment size on the test set using calibrated LateFusion predictions. Relative risks are reported with respect to the baseline prevalence within each enrollment group.
}
\label{tab:risk_stratification_by_enrollment}
\end{table}

\end{document}